%% file: camera_ready.tex
\pdfoutput=1
\documentclass{article}

\usepackage{microtype}
\usepackage{graphicx}
\usepackage{subcaption}
\usepackage{booktabs} %
\usepackage{pgfplots}
\usepackage{pgfplotstable}
\definecolor{nicepurple}{HTML}{8859AB}
\definecolor{nicegreen}{HTML}{88C2AB}
\definecolor{niceorange}{HTML}{CB722F}
\definecolor{niceblue}{HTML}{287DC5}
\definecolor{nicegray}{HTML}{363635}

\pgfplotsset{compat=1.14}

\pgfplotstableset{
    mean pm std/.style args={#1 from #2 with #3}{
        create on use/#1/.style={
            create col/assign/.code={
                \edef\mean{\thisrow{#2}}%
                \edef\std{\thisrow{#3}}%
                \def\dash{--}%
                \ifx\mean\dash
                    \pgfkeyssetvalue{/pgfplots/table/create col/next content}{--}%
                \else
                    \ifx\std\dash
                        \edef\temp{\noexpand\pgfmathprintnumber[fixed, fixed zerofill, precision=1]{\mean}}%
                        \pgfkeyslet{/pgfplots/table/create col/next content}{\temp}%
                    \else
                        \edef\temp{\noexpand\pgfmathprintnumber[fixed, fixed zerofill, precision=1]{\mean} {\noexpand\scriptsize $\noexpand\pm$ \noexpand\pgfmathprintnumber[fixed, fixed zerofill, precision=1]{\std}}}%
                        \pgfkeyslet{/pgfplots/table/create col/next content}{\temp}%
                    \fi
                \fi
            }
        }
    }
}

\usepackage{hyperref}

\usepackage[accepted]{icml2026}

\usepackage{amsmath}
\usepackage{amssymb}
\usepackage{mathtools}
\usepackage{amsthm}
\usepackage{bm}

\usepackage{hyperref}
\usepackage{fontawesome5}
\usepackage{xcolor}
\usepackage[most]{tcolorbox}

\usepackage{pgfplots}
\pgfplotsset{compat=1.18}

\usepackage[capitalize,noabbrev]{cleveref}

\theoremstyle{plain}

\theoremstyle{definition}

\theoremstyle{remark}

\usepackage[textsize=tiny]{todonotes}

\icmltitlerunning{Efficient Edge-aware Attention Network for Graph Generation}

\begin{document}

\twocolumn[
  \icmltitle{Efficient Edge-aware Attention Network for Graph Generation}

  \icmlsetsymbol{equal}{*}

  \begin{icmlauthorlist}
    \icmlauthor{Vincent Jung}{equal,yyy,sch}
    \icmlauthor{Alba Carballo-Castro}{equal,yyy}
    \icmlauthor{Yiming Qin}{equal,yyy}
    \icmlauthor{Lonneke van der Plas}{sch}
    \icmlauthor{Pascal Frossard}{yyy}
  \end{icmlauthorlist}

  \icmlaffiliation{yyy}{LTS4, EPFL, Lausanne, Switzerland}
  \icmlaffiliation{sch}{Università della Svizzera italiana, Lugano, Switzerland}

  \icmlcorrespondingauthor{}{first\_name.last\_name@epfl.ch}

  \icmlkeywords{Machine Learning, ICML}

  \vskip 0.3in
]

\printAffiliationsAndNotice{\icmlEqualContribution}

\begin{abstract}
Graph generative models commonly use graph-specific Transformer architectures to jointly update node and edge features. However, these architectures do not directly benefit from recent hardware-aware attention implementations, 
limiting their scalability. The adaptation is, however, non-trivial: edge and graph-level features must be injected into pairwise attention scores, graph masks must be respected, and dynamic edge updates remain a major source of quadratic memory usage. We propose an edge-aware attention layer for graph generation that incorporates structural information through head-wise score modulation, while updating edge representations with a lightweight residual mechanism that retains attention-score dependent edge updates. This layer can be effectively implemented using PyTorch's \texttt{FlexAttention}. We achieve similar generation quality to our baseline while reducing peak GPU memory by around 65\% and substantially improving training and sampling throughput. These results suggest that the proposed architecture is a practical step toward scalable graph foundation models.

\end{abstract}

\vspace{-0.5cm}
\begin{center}
\href{https://github.com/vjung99/flexattention-graph-transformer}{%
  \tcbox[
    enhanced,
    on line,
    colback=gray!10,
    colframe=gray!40,
    arc=3.5mm,
    boxrule=1pt,
    top=2pt,bottom=2pt,
    left=6pt,right=6pt,
    coltext=darkgray
  ]{\small\faGithub\ \textbf{Code}}%
}
\end{center}
\vspace{-0.1cm}

\section{Introduction}
Graph generation~\citep{simonovsky2018graphvae,martinkus2022SPECTRE, jo2024GruM} has advanced substantially with discrete diffusion and flow-based models. A particularly influential line of work, initiated by the diffusion model DiGress~\citep{vignac2022digress}, and then flow-based models such as CatFlow~\citep{Eijkelboom2024VariationalFM}, DeFoG~\citep{qin2025defog}, BWFlow~\citep{jiang2025bures}, and VBFN~\citep{xiong2026variationalbayesianflownetwork}, and diffusion models such as Cometh~\citep{siraudin2024cometh}, relies on a specific transformer backbone to denoise categorical node and edge features. This architecture is inspired by the original Graph Transformer~\cite{dwivedi2021graphtransformer}, which already updates edge states from attention scores and uses edge states to modulate self-attention. The main differences are that it computes attention between all nodes as opposed to restricting it to neighbors, maintains a graph-level feature vector and uses FiLM \citep{perez2018film} modules to couple node, edge, and graph-level features. This dense, layered interaction is central to its success in graph generation, where the model must capture both local connectivity and global graph structure.

However, this expressivity comes with a computational cost. Since the model must maintain dense pairwise edge states, both attention computation and dynamic edge representations scale quadratically with the number of nodes, making training expensive and difficult under limited GPU resources. A large body of work addresses this bottleneck through sparse attention patterns~\citep{shirzad2023exphormer}; hybrid local-global designs~\citep{rampasek2022gps}; linear and operator-based alternatives~\citep{deng2024polynormer,sancak2024geco}; or tokenization-based formulations that cast graphs as sequences of node and edge tokens and apply standard Transformers with minimal graph-specific modifications~\cite{kim2022puretransformerspowerfulgraph}. These methods are effective, but typically modify or replace dense self-attention altogether. In graph generation specifically, SparseDiff~\citep{qin2025sparsediff} exploits sparsity to enable training on larger graphs, although this increases sampling time. 

In this work, we take a complementary route and ask whether the original dense edge-aware architecture can be implemented more efficiently while fully preserving its inductive bias. We focus on the computationally-expensive and memory-hungry attention layer in the original implementation, and design a kernel-friendly version which still modulates attention scores using edge and global features, respects graph masks, and enables edge features update from its attention map. This version can be implemented entirely using \texttt{FlexAttention}~\citep{dong2025flexattention}, which supports custom attention score modifications and masking. This enables us to use fused attention kernels, bringing throughput and memory gains. 

Empirically, our proposed model (FlexAttentionGT) matches the generation quality of DeFoG~\cite{qin2025defog}'s original backbone while reducing peak memory by 65\% and improving training and inference throughput by 45\% and 94\% on average. Ablations show that the edge update is critical: simplifying it causes generation collapse, whereas reducing model size leads to slower convergence and worse final performance. Together, these results demonstrate that edge-aware graph Transformers can be made more efficient with tailored design while preserving their expressivity, supporting their scalability to the larger graphs encountered in foundation model settings.

\section{Background and related work}

\paragraph{Generative Graph Transformer}
Recent works in graph generation such as DiGress~\citep{vignac2022digress} or BWFlow~\cite{jiang2025bures} employ a Graph Transformer (GT) backbone that jointly updates node features $\bm{X}\in\mathbb{R}^{N\times d_x}$, edge features $\bm{E}\in\mathbb{R}^{N\times N\times d_e}$, and graph-level features $\bm{y}\in\mathbb{R}^{d_y}$, with $N$ the number of nodes of the largest graph in a batch. In contrast to the original Graph Transformer~\cite{dwivedi2021graphtransformer}, this backbone keeps dense edge features and graph-level features, and injects them into node attention through FiLM-style score modulation~\citep{perez2018film} (Further details in Appendix \ref{app:original}). The attention map is then reused to update edge features, while graph-level features are updated from pooled node and edge representations.
This design is effective for graph generation because it tightly couples node-, edge-, and graph-level information. However, it bottlenecks efficiency: the model maintains dense pairwise edge states $\bm{E}$, which need to be included in the node attention, and themselves updated using these attention scores, resulting in quadratic computational and memory costs per layer.

\paragraph{Structural attention biases and edge-aware architectures.}
A central approach to injecting graph structure into Transformers is modulating attention logits with structural features. Graphormer~\citep{ying2021graphormer} introduced spatial and edge encodings as additive biases on pre-softmax logits. EGT~\citep{hussain2022egt} proposed bidirectional coupling between explicit edge channels and node attention, architecturally close to the generative GT backbone we build upon. Relational Attention~\citep{diao2023relationalattention} similarly updates edge vectors at each layer. Our work preserves this dense edge coupling while making it compatible with fused attention kernels.

\paragraph{Adjacency masking and hybrid designs.}
Exphormer~\citep{shirzad2023exphormer} builds sparse attention from local neighborhoods, expander graphs and virtual global nodes, yielding a scalable graph transformer; GraphGPS~\citep{rampasek2022gps} runs a local MPNN and a global Transformer in parallel, avoiding committing entirely to dense or sparse attention. In graph generation, SparseDiff~\citep{qin2025sparsediff} exploits graph sparsity to scale to larger graphs, but relies on message-passing layers that are less amenable to efficient parallel computation.

\paragraph{Scalable alternatives.}
Polynormer~\citep{deng2024polynormer} combines linear-time attention with a polynomial-expressive formulation to balance scalability and expressivity, GECO~\citep{sancak2024geco} proposes a scalable operator-based alternative, and tokenization-based methods~\citep{kim2022puretransformerspowerfulgraph} apply standard Transformers to node and edge token sequences with minimal graph-specific modifications. These approaches are effective, but typically modify or replace dense attention altogether.

\paragraph{Graph foundation models.}
As graph learning scales toward foundation model settings, the cost of dense edge-aware attention becomes increasingly prohibitive. GraphBFF~\citep{bechlerspeicher2026billionscalegraphfoundationmodels} addresses this for large heterogeneous graphs by replacing dense attention with type-aware sparse attention. Our work takes a complementary route: rather than replacing the dense backbone, we make it hardware-efficient through fused kernels, preserving the inductive bias that drives generation quality and enabling its application at larger scales.

We refer the reader to Appendix~\ref{app:relatedworks} for a more detailed overview of the relevant literature.

\section{Efficient Attention Layers for Graph Generation}

We identify three challenges in making the generative GT backbone compatible with fused attention kernels. (i)~The original architecture uses element-wise outer-product attention, producing a $d_f$-dimensional tensor per node pair that standard fused kernels cannot process. (ii)~Edge and global features must still modulate pairwise attention scores in a way that is both expressive and kernel-compatible. (iii)~Edge representations depend on the softmax attention probabilities $\alpha$, which fused kernels do not materialize in global memory. We address each challenge in turn, with further implementation details in Appendices~\ref{app:flexatt} and~\ref{app:flexatt_details}.

\paragraph{\texttt{FlexAttention}.}
We leverage \texttt{FlexAttention} \citep{dong2025flexattention}, a PyTorch interface that lowers customized attention variants to fused Triton kernels. It exposes two hooks: \texttt{score\_mod} modifies each pre-softmax attention logit, enabling structural biases and edge-dependent score modulation; \texttt{mask\_mod} specifies which query-key pairs are valid, compiled into a block-sparse \texttt{BlockMask} that allows the kernel to skip fully-masked blocks. This interface is well-suited for injecting graph masks and structural biases into attention; however, it does not natively support the edge update used in generative graph Transformers, since it only returns the attended node values. We address this below.

\paragraph{From element-wise to scalar dot-product attention.}
The original backbone computes an unnormalized attention tensor $\bm{Y}\in \mathbb{R}^{N \times N \times H \times d_f}$ via the element-wise outer product $\bm{Y}_{ij,h} = \bm{q}_{i,h} \odot \bm{k}_{j,h}$, retaining a $d_f$-dimensional feature vector per node pair and head. Softmax is then applied \emph{independently per channel}, yielding $d_f$ separate attention distributions per query-head pair. While expressive, this formulation cannot be lowered to a standard fused kernel. We replace it with standard scaled dot-product attention. The unnormalized scalar logit for each head $h$ is 
\begin{equation}
    s_{ij,h} = \frac{\bm{q}_{i,h} \cdot \bm{k}_{j,h}}{\sqrt{d_f}} \in \mathbb{R},
\end{equation}
giving a single attention distribution per query-head pair and making the node-update compatible with standard FlexAttention.

\paragraph{Head-wise structural FiLM layer}
To inject edge features $\bm{E} \in \mathbb{R}^{N \times N \times d_e}$ and global features $\bm{y} \in \mathbb{R}^{d_y}$ into attention, we learn one scalar per edge and attention head:
\begin{equation}
    s'_{ij,h} = s_{ij,h} \cdot \bigl(E^{\mathrm{mul}}_{ij,h} + 1\bigr) + E^{\mathrm{add}}_{ij,h},
\end{equation}
where $\bm{E}^{\mathrm{mul}}, \bm{E}^{\mathrm{add}} \in \mathbb{R}^{N \times N \times H}$ are obtained by linearly projecting $E$ and $y$ to $H$-dimensional vectors. This head-wise FiLM-style modulation is applied \emph{inside} the fused kernel via the \texttt{score\_mod} hook, so no quadratic tensor is written to global memory during the attention pass. Padding and graph masks are handled via a \texttt{BlockMask}, allowing the kernel to skip fully-masked blocks.

\paragraph{Single-pass node and edge update via $\bm{I_N}$ concatenation.}
A key challenge is that edge representations in the original backbone are updated from pre-softmax attention scores, which FlexAttention does not return. Instead, we use the normalized attention coefficients $\bm{\alpha} \in \mathbb{R}^{N\times N\times H}$ and recover them in a \emph{single} FlexAttention call by concatenating an identity matrix $\bm{I_N} \in \mathbb{R}^{N \times N}$ to the value matrix:
\begin{equation}
    \label{eq:v_aug}
    \mathrm{FlexAttn}(\bm{Q},\, \bm{K},\, [\bm{V} \,\|\, \bm{I_N}])
    \;=\; \bm{\alpha}[\bm{V} \,\|\, \bm{I_N}]
    \;=\; [\bm{\alpha} \bm{V} \,\|\, \bm{\alpha}].
\end{equation}

The kernel simultaneously produces the aggregated node values $\bm{V'} = \bm{\alpha} \bm{V} \in \mathbb{R}^{N \times H \times d_f}$ and the attention maps $\bm{\alpha} \in \mathbb{R}^{N \times H\times N}$ as the trailing output channels. Edge representations are updated using:
\begin{equation}
    \label{eq:edge_update}
    E'_{ij} = \mathrm{Linear}\!\left(\alpha_{ij} \odot \phi(\bm{y})\right) + E_{ij},
\end{equation}
where $\phi(\bm{y})$ provides a FiLM scale and bias learned from the global features $\bm{y}$. Node and edge updates thus share a single function call. This works well for the benchmark datasets we consider here, which contain graphs with fewer than 256 nodes. For larger graphs, the concatenation of $N$ channels to the value vectors increases the SRAM requirements, which necessitates a smaller block size and leads to worse throughput. In these cases, we calculate the attention output and the attention map in two separate passes, which allows us to keep an appropriate block size. We address this in more details in Appendix \ref{app:efficiency_larger_graphs}.

The edge update remains quadratic in $N$, but the dominant constant is reduced because the heavy attention computation itself is handled by FlexAttention.

\section{Experiments}
\label{sec:experiments}

We train our FlexAttentionGT architecture and the original GT backbone using the DeFoG~\citep{qin2025defog} framework, and evaluate on the Planar, SBM, and Tree datasets originally used in DeFoG, while keeping all training hyperparameters fixed across configurations. Performance is assessed using validity, uniqueness, and novelty (V.U.N.~$\uparrow$), as well as the Ratio~$\downarrow$ metric comparing graph-statistic distances between generated and test graphs relative to the train--test baseline.

\begin{figure*}[ht]
    \centering
    \captionsetup[subfigure]{
        labelfont=normalfont,
        textfont=normalfont,
        justification=centering,
        margin={10mm,0mm}
    }
    \begin{subfigure}[b]{0.32\textwidth}
        \centering
        \input{plot_inference.tex}
        \caption{Inference Speed}
    \end{subfigure}
    \begin{subfigure}[b]{0.32\textwidth}
        \centering
        \input{plot_training.tex}
        \caption{Training Speed}
    \end{subfigure}
    \begin{subfigure}[b]{0.32\textwidth}
        \centering
        \input{plot_memory.tex}
        \caption{Memory Consumption}
    \end{subfigure}
    
    \vspace{0.1cm}
    \ref*{sharedlegend}
    \caption{Computational efficiency across datasets. Percentages denote the relative change compared to the original implementation.}
    \label{fig:efficiency_barplots}
\end{figure*}
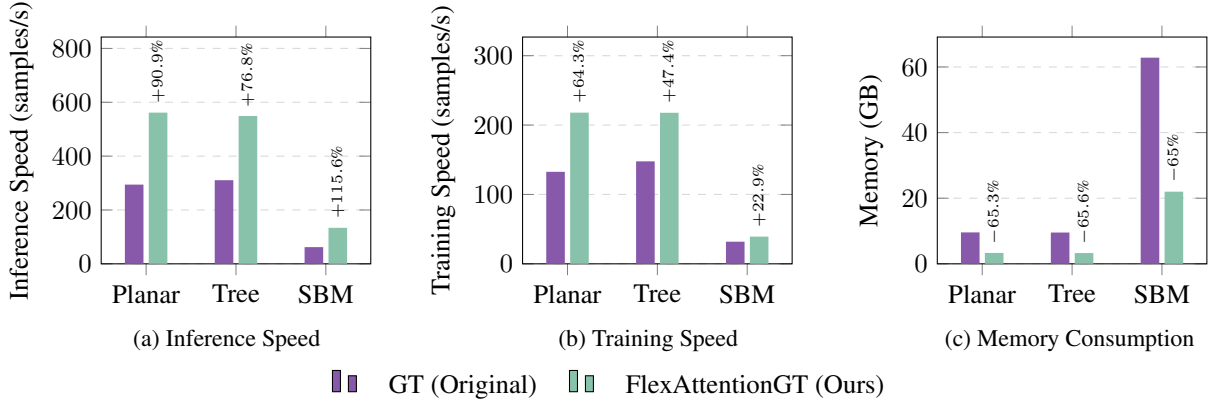

\subsection{Graph generation performance}

Table~\ref{tab:generation_results} reports generation quality across all datasets. Our updated architecture matches or slightly improves upon the original GT on V.U.N.\ metrics, including a small gain on SBM. Although the Ratio metric is marginally higher on the SBM dataset, the differences remain within one standard deviation, which indicates comparable structural fidelity overall.

\begin{table}[ht]
\caption{Generation performance on Planar, Tree, and SBM.}
\label{tab:generation_results}
\centering
\begin{tabular}{lcc}
\toprule
Model & V.U.N.~$\uparrow$ & Ratio~$\downarrow$ \\
\midrule
\multicolumn{3}{c}{\textit{Planar}} \\
\midrule
Train set             & $100$              & $1.00$           \\
GT (Original)      & $99.5 \pm 1.0$     & $2.47 \pm 0.33$  \\
FlexAttentionGT (Ours) & \bm{$100.0 \pm 0.0$} & \bm{$2.21 \pm 0.95$ } \\
\midrule
\multicolumn{3}{c}{\textit{Tree}} \\
\midrule
Train set             & $100$              & $1.00$           \\
GT (Original)      & \bm{$97.0 \pm 3.7$}     & $2.24 \pm 0.63$  \\
FlexAttentionGT (Ours) & \bm{$97.0 \pm 1.9$}    & \bm{$2.12 \pm 0.87$}  \\
\midrule
\multicolumn{3}{c}{\textit{SBM}} \\
\midrule
Train set             & $85.9$             & $1.00$           \\
GT (Original)      & $85.0 \pm 5.3$     & \bm{$3.39 \pm 1.08$}  \\
FlexAttentionGT (Ours) & \bm{$89.5 \pm 5.7$}     & $3.70 \pm 1.08$  \\
\bottomrule
\end{tabular}
\end{table}

\begin{figure*}[t]
    \centering
    \captionsetup[subfigure]{
        labelfont=normalfont,
        textfont=normalfont,
        justification=centering,
        margin={8.5mm,0mm}
    }
    \begin{subfigure}{0.245\textwidth}
        \centering
        \input{plot_curve1000_planar.tex}
        \caption{Planar: 1000 gen. steps}
        \label{fig:planar_1000}
    \end{subfigure}
    \hfill
    \begin{subfigure}{0.245\textwidth}
        \centering
        \input{plot_curve1000_sbm.tex}
        \caption{SBM: 1000 gen. steps}
        \label{fig:sbm_1000}
    \end{subfigure}
    \hfill
    \begin{subfigure}{0.245\textwidth}
        \centering
        \input{plot_curve50_planar}
        \caption{Planar: 50 gen. steps}
        \label{fig:planar_50}
    \end{subfigure}
    \begin{subfigure}{0.245\textwidth}
        \centering
        \input{plot_curve50_sbm}
        \caption{SBM: 50 gen. steps}
        \label{fig:sbm_50}
    \end{subfigure}
    \vspace{0.2cm}
    
    \ref*{sharedlegend}
    \caption{V.U.N. curves over training steps on the Planar and SBM datasets, comparing FlexAttentionGT against the original GT and its lightweight variants. We evaluate 1000 and 50 generation steps, respectively.}
    \label{fig:gen_quality_curves}
\end{figure*}
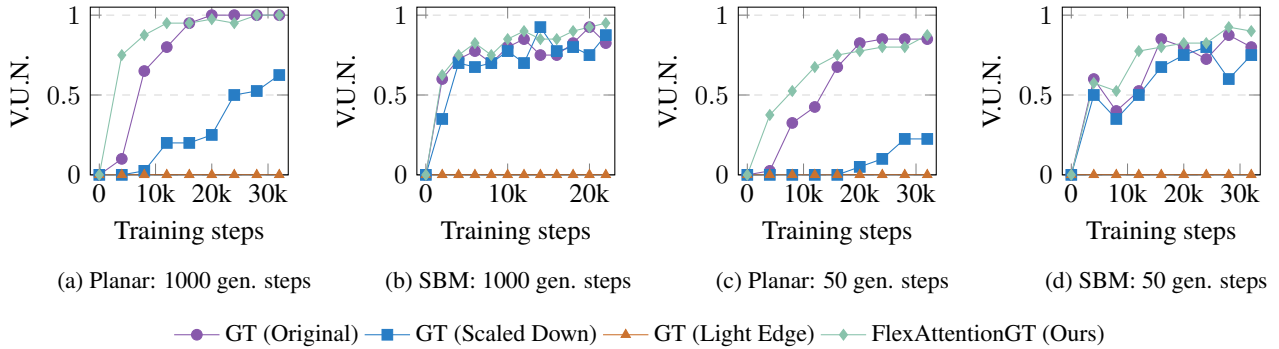

\subsection{Speedup and memory usage improvement}

Figure~\ref{fig:efficiency_barplots} compares inference throughput, training speed, and peak GPU memory across datasets. Our model is substantially faster than the original one on both training and inference, with speedups ranging from 23\% to 116\% depending on the dataset and regime. Peak memory is reduced by approximately 65\% across all datasets, roughly one third of the original. These gains are consistent across datasets, indicating that the efficiency improvements are not dataset-specific.

\subsection{Additional experiments}

To isolate the contribution of our key design choices, we explore the effect of the number of generation steps (1000 vs.\ 50), model size via a scaled-down version of the original implementation (\textbf{Scaled Down}), and the edge update strategy via a variant without per-edge modulation of attention scores (\textbf{Light Edge}), both matched to our architecture in peak memory.

\paragraph{Generation quality across training with 1000 steps.}

Figures~\ref{fig:planar_1000} and \ref{fig:sbm_1000} show V.U.N.\ curves on Planar and SBM with 1000 generation steps. Our model matches or improves on all baselines on convergence and final performance. GT (Scaled Down) learns more slowly and underperforms, showing that matching memory by reducing capacity is a poor trade-off. GT (Light Edge) fails to learn entirely, confirming that per-edge modulation of attention scores is a critical component. The effect is most pronounced on Planar, but our model also converges fastest on SBM.

\paragraph{Generation quality across training with 50 steps.}
Figures~\ref{fig:planar_50} and \ref{fig:sbm_50} repeat the same comparison under a reduced budget of 50 generation steps. Under this constrained setting, GT (Light Edge) again fails to learn, and GT (Scaled Down) also fails in the Planar dataset. The original GT shows moderate but inconsistent performance, while our model achieves the best and most stable results across training, demonstrating improved sample efficiency and robustness in the low-step regime. 

\begin{figure}[htb]
    \centering
    \includegraphics[width=1.0\linewidth]{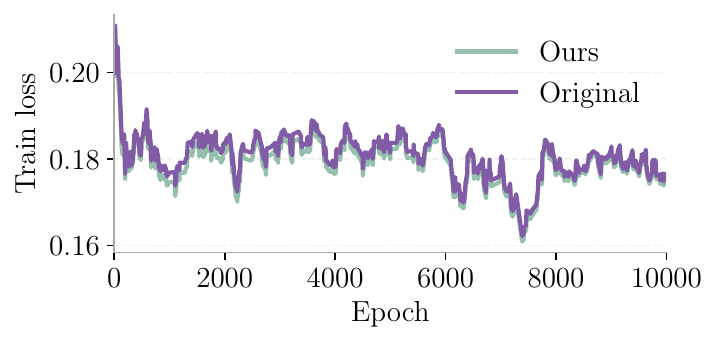}
    \caption{Training loss comparison between FlexAttentionGT (Ours) and GT (Original), smoothed with a window size of 20. Similar optimization dynamics suggest that gains stem from the attention layer's inductive bias.}
    \label{fig:training_loss_comparison}
\end{figure}

\paragraph{Training dynamics.}
Figure~\ref{fig:training_loss_comparison} shows training loss curves on Planar for the original and updated models. The two curves overlap closely throughout training, indicating that the architectural changes do not alter the optimization landscape. This suggests that the observed quality differences are unlikely to originate from a change in how the model optimizes, but rather a benefit of the inductive biases of our attention layer.

\begin{figure}[ht]
    \centering
    \input{plot_curve1000_planar_ablation_attncoeff.tex}
    \ref*{sharedlegend_ablation_edge}
    \caption{V.U.N. curves comparing normalized and unnormalized networks over training steps on the Planar dataset, evaluated with 1000 generation steps.}
    \label{fig:ablation_edge_update}
\end{figure}
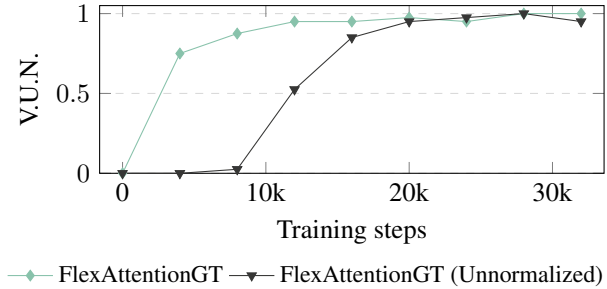

\paragraph{Edge update using normalized attention coefficients.}
Our edge updates use the normalized attention coefficients $\alpha_{ij,h}$ recovered using (Eq.~\ref{eq:v_aug}), whereas the baseline implementation uses unnormalized attention scores. To isolate the effect of this change, we implement a version of FlexAttentionGT which separately calculates the unnormalized attention coefficients and uses those for edge updates. We train this version on the planar dataset and report the V.U.N. curves with 1000 generation steps for both implementations in Figure~\ref{fig:ablation_edge_update}. The unnormalized variant converges more slowly, showing that the normalization provided by softmax is beneficial for stable edge feature learning.

\section{Conclusion}

In this work, we presented an efficient and expressive edge-aware graph Transformer that leverages recent fused attention kernels for graph generation. By shifting from element- to head-wise FiLM transformations, we eliminated major memory bottlenecks while preserving nuanced structural bias injection. Furthermore, our approach allows us to seamlessly extract the attention map and concurrently update edge features within a single kernel pass.

Our empirical benchmarks across Planar, Tree, and SBM datasets demonstrate that this architectural refinement yields an average of $\bm{44.8\%}$ speedup in training, an average of $\bm{94.4\%}$ speedup in inference, and consumes $\bm{65\%}$ less peak GPU memory compared to the original dense implementation, with no loss in generation fidelity. These gains can be particularly relevant in the context of graph foundation models, where scaling edge-aware architectures to larger and more diverse graphs is a key challenge.

Beyond graph generation, the design principles introduced here (head-wise structural conditioning inside a fused kernel and single-pass node-edge updates) could be applicable to any edge-aware transformer setting, including those used in graph classification, regression, and link prediction, offering a practical path toward more scalable graph learning.

\newpage

\section*{Acknowledgements}
This project was supported by the Swiss National Science Foundation (SNSF grant 10001445). Vincent Jung is partially supported by the Swiss National Science Foundation (SNSF grant 205121 207437: C-LING) and the Fondazione Aldo e Cele Daccò.

\section*{Use of Large Language Models}

Large Language Models (LLMs) were used as a coding and writing assistant tool in the preparation of this manuscript and its experiments. Specifically, they were employed to aid in writing, checking spelling, and editing for clarity. They were also used to assist in coding and plotting of the results. Nonetheless, all text and code produced with the assistance of LLMs was carefully reviewed, verified, and revised by the authors to ensure accuracy and appropriateness. The use of LLMs was limited to these support functions, and the authors take full responsibility for the final contents.

\section*{Impact Statement}

The primary objective of this paper is to advance graph generation with a more efficient network. While this development has the potential to bring about both positive and negative societal or ethical impacts, particularly in areas like biomedical and chemical research, we currently do not foresee any immediate societal concerns associated with the proposed methodology.

{\interlinepenalty=10000
\bibliography{camera_ready}
\bibliographystyle{icml2026}
}

\newpage
\appendix
\onecolumn
\section{Extended Related Work}
\label{app:relatedworks}
 
In the following, we take a deeper look and position our work within the broader literature on Graph Transformers. We cover structural attention biases, attention masking, efficient attention mechanisms, and more recent graph transformer architectures motivated by graph foundation models. We refer the reader to recent surveys for a more exhaustive and complementary overview
\citep{shehzad2024graphtransformers,yuan2025surveygraphtransformers}.
 
\subsection{Graph Transformers with Structural Attention Biases}
 
A central challenge in applying Transformers to graphs is injecting graph-structural information into the attention computation. Graphormer~\citep{ying2021graphormer} introduced centrality, spatial, and edge encodings that modulate pre-softmax logits and node input features, enabling standard dot-product attention to be directly conditioned on shortest-path distances and edge features, showing that well-designed structural biases suffice to bring a vanilla Transformer to state-of-the-art performance on graph-level tasks. Edge-augmented Graph Transformer~(EGT; \citealt{hussain2022egt}) proposed explicit residual \emph{edge channels} that are bidirectionally coupled with node attention: edge embeddings bias the attention weights, and the resulting pre-softmax scores in turn update the edge representations. This bidirectional coupling is architecturally close to the generative GT backbone built on~\citep{dwivedi2021graphtransformer}, the latter differing mainly in its use of FiLM-style modulation and the joint maintenance of graph-level features. Relational Attention~\citep{diao2023relationalattention} generalizes Transformer attention to jointly consider and update edge vectors at each layer, achieving strong results on algorithmic reasoning benchmarks. Finally, the Generalized-Distance Transformer~(GDT; \citealt{stoll2026generalizable}) offers a unified analysis of structural encodings and attention mechanisms across a large evaluation suite, distilling design choices that consistently transfer across tasks and scales.
 
Our work shares the same spirit of preserving dense edge representations, but focuses instead on making the resulting architecture compatible with hardware-fused attention kernels without sacrificing expressivity.
 
\subsection{Graph Transformers with Adjacency Masking}
 
An alternative to score-level structural biases is to \emph{mask} the attention matrix so that each node attends only to a structured neighborhood. Exphormer~\citep{shirzad2023exphormer} builds a sparse attention graph from expander edges, virtual global nodes and local neighborhood edges, achieving attention cost linear in the number of nodes and edges. GraphGPS~\citep{rampasek2022gps} introduced a hybrid architecture that runs a local MPNN and a global Transformer in parallel, adding their outputs at each layer; this recipe has become a strong baseline for graph learning precisely because it avoids committing entirely to either sparse or dense attention. Edge-set Attention~(\citealt{buterez_end--end_2025}) explores graphs as edge sets and vertically interleaves masked and vanilla self-attention modules, demonstrating that simple connectivity-based masking can match or exceed more complex graph transformers, including on long-range benchmarks. More recently, DAM-GT~\citep{li2025damgt} introduced a masking scheme and an attribute-aware dual positional encoding, specifically designed to prevent attention diversion in neighborhood-token architectures for node classification.
 
\subsection{Scalable and Efficient Attention for Graphs}
 
Reducing the quadratic computational and memory cost associated with dense attention has become a major focus of research in recent years, tackled from several complementary angles.
 
\paragraph{Linear and operator-based alternatives.}
Polynormer~\citep{deng2024polynormer} combines linear-time attention with a polynomial-expressive formulation to balance scalability and expressivity, while GECO~\citep{sancak2024geco} proposes a scalable operator-based alternative to full graph Transformers. DUALFormer~\citep{zhuo2025dualformer} introduces a dual-dimensional design that performs self-attention in the feature dimension rather than the node dimension, achieving scalability with linear-complexity attention. UGCFormer~\citep{zhuo2026closer} identifies a shared \emph{cross-aggregation} mechanism underlying existing Graph Transformers and proposes a linearized dual cross-attention module (DCA) to realize it. Powerful Plain Graph Transformers (PPGT,~\citealt{ma2026plain}) demonstrates that three minimal modifications to the standard Transformer ($L_2$ attention, adaptive RMS normalization, and an MLP-based stem for positional encoding) suffice to achieve competitive graph learning without any message-passing components, narrowing the gap between general-purpose and graph-specific architectures.
 
\paragraph{Sparse implementations.}
\citet{tomczak2025sparsegraphprocessing} treat sparse attention masks as graph adjacency patterns and develop graph processing algorithms to obtain work-optimal attention implementations; their framework complements our fused-kernel approach by targeting scenarios where true sparsity can be exploited. In the graph-generation setting specifically, SparseDiff~\citep{qin2025sparsediff} exploits graph sparsity to scale discrete diffusion to larger graphs, but relies on message-passing layers which are less easily parallelizable. FlexGNN~\citep{bae2025flexgnn} addresses full-graph GNN training at scale by managing intermediate data and optimizing inter-GPU communication plans, and is orthogonal to our work on attention kernel efficiency.
 
\paragraph{Towards graph foundation models.}
GraphBFF~\citep{bechlerspeicher2026billionscalegraphfoundationmodels} is an end-to-end recipe for
billion-parameter Graph Foundation Models on heterogeneous graphs. In particular, they present the GraphBFF transformer, which replaces dense attention with type-aware sparse attention and degree-constrained neighbor sampling, enabling scalable learning on highly heterogeneous, large-scale graphs. The need for these efficiency mechanisms highlights the importance of reducing the memory cost of dense, edge-aware graph architectures.
 
\subsection{Architecture Search and Tokenization-Based Methods}
 
AutoGT~\citep{zhang2023autogt} proposes an automated search over the joint space of graph encoding strategies and Transformer architectures, unifying most existing designs into a single searchable framework. \citet{kim2022puretransformerspowerfulgraph} reformulate graphs as sequences of node and edge tokens and apply a standard Transformer with minimal graph-specific modifications, highlighting that carefully designed tokenization can compensate for the absence of explicit structural biases. The GDT analysis~\citep{stoll2026generalizable} also supports edge-level tokenization.
 
\section{Efficiency Benchmarks}

\subsection{Detailed efficiency benchmarks on Planar, Tree and SBM}
In this section, we provide the exact numerical values corresponding to the barplots presented in Section \ref{sec:experiments}. Table \ref{tab:efficiency_detailed} summarizes the inference speed, training speed, and peak GPU memory consumption during training for both implementations across the evaluated datasets.

\input{table_efficiency.tex}

\subsection{Efficiency benchmark on larger graphs}
\label{app:efficiency_larger_graphs}
The $I_N$ concatenation trick used in our single-pass edge update strategy increases the SRAM requirements for each block in the FlexAttention kernel. For large graphs above 256 nodes, this forces us to reduce the block size leading to a slowdown because of the larger number of kernel calls. We implement a two-pass version of our architecture where the attention output and the attention map are calculated in two separate passes, allowing us to keep a larger block size. We compare the throughput and memory requirement of this implementation with the single-pass version on the SBM dataset generated with graphs of size 400 to 600 nodes. The results are shown in Table \ref{tab:efficiency_large_graphs}. We observe that the two-pass version achieves higher throughput than the single-pass version while being mathematically equivalent, and still delivering large gains compared to the original implementation.

Future work can explore custom kernels which perform the update of the attention map using input edge features and the subsequent edge feature updates using said attention map in a block-wise manner, removing the need to extract the entire attention map to global memory and thus preserving the linear memory scaling of FlexAttention even with the edge update. This should be feasible since the inclusion of edge features and the edge update are both pointwise operations that can be performed for a pair of nodes (i,j) independently of other pairs, at least when using unnormalized attention coefficients for the edge updates. This could not only reduce memory requirements due to the attention calculation itself, but also reduce the amount of intermediate edge features that need to be stored in global memory, which become the major bottleneck for large graphs.

\input{table_efficiency_large_graphs.tex}

\section{Graph Transformer Layer Details}
\label{app:transformer}

We describe the two layer variants in full detail. Throughout, let $N$ denote the number of nodes, $H$ the number of attention heads, and $d_f = d_x / H$ the per-head feature dimension. We write $X \in \mathbb{R}^{N \times d_x}$, $E \in \mathbb{R}^{N \times N \times d_e}$, and $y \in \mathbb{R}^{d_y}$ for the node, edge, and global feature matrices respectively. All linear projections are written as $\mathrm{Linear}(\cdot)$ with output dimensions indicated in context. We use $\odot$ for element-wise multiplication and $[\,\cdot \,\|\, \cdot\,]$ for concatenation along the last axis.

\subsection{Original Layer}
\label{app:original}

\paragraph{Query, key, and value projections.}
Node features are projected to queries, keys, and values:
\begin{equation}
    Q = \mathrm{Linear}_{d_x}(X), \quad
    K = \mathrm{Linear}_{d_x}(X), \quad
    V = \mathrm{Linear}_{d_x}(X),
\end{equation}
each of shape $\mathbb{R}^{N \times d_x}$, then reshaped to $\mathbb{R}^{N \times H \times d_f}$.

\paragraph{Element-wise outer-product attention.}
Rather than computing a scalar dot product, the original layer forms a $d_f$-dimensional bilinear tensor per edge:
\begin{equation}
    Y_{ij,h} = \frac{q_{i,h} \odot k_{j,h}}{\sqrt{d_f}} \in \mathbb{R}^{d_f},
\end{equation}
yielding $Y \in \mathbb{R}^{N \times N \times H \times d_f}$. This retains the full per-channel interaction between $q_{i,h}$ and $k_{j,h}$.

\paragraph{Edge FiLM on attention scores.}
Edge features are projected to per-channel multiplicative and additive modifiers:
\begin{equation}
    E^{\mathrm{mul}} = \mathrm{Linear}_{d_x}(E), \quad
    E^{\mathrm{add}} = \mathrm{Linear}_{d_x}(E),
\end{equation}
both reshaped to $\mathbb{R}^{N \times N \times H \times d_f}$, and applied as a FiLM transform~\cite{perez2018film} over the full feature dimension:
\begin{equation}
    Y_{ij,h} \leftarrow Y_{ij,h} \odot \bigl(E^{\mathrm{mul}}_{ij,h} + 1\bigr) + E^{\mathrm{add}}_{ij,h}.
\end{equation}

\paragraph{Edge update.}
The modulated tensor $Y$ is flattened to $\mathbb{R}^{N \times N \times d_x}$ and conditioned on global features via a second FiLM transform:
\begin{equation}
    \hat{E}_{ij} = \mathrm{Linear}_{d_y}(y) + \bigl(\mathrm{Linear}_{d_y}(y) + 1\bigr) \odot Y^{\mathrm{flat}}_{ij},
\end{equation}
followed by a final linear projection to $\mathbb{R}^{d_e}$.

\paragraph{Per-channel softmax and node update.}
Softmax is applied over the key dimension independently for each channel:
\begin{equation}
    \alpha_{ij,h,c} = \frac{\exp(Y_{ij,h,c})}{\sum_{j'} \exp(Y_{ij',h,c})},
    \quad c = 1,\ldots,d_f,
\end{equation}
yielding $d_f$ independent attention distributions per query-head pair.
The weighted sum over values is then:
\begin{equation}
    \hat{x}_{i,h,c} = \sum_j \alpha_{ij,h,c} \cdot v_{j,h,c},
\end{equation}
and the result is flattened to $\mathbb{R}^{N \times d_x}$.

\paragraph{Global FiLM on node update and node output.}
Global features modulate the aggregated node representations before the output projection:
\begin{equation}
    \hat{X} \leftarrow \mathrm{Linear}_{d_y}(y) + \bigl(\mathrm{Linear}_{d_y}(y) + 1\bigr) \odot \hat{X}.
\end{equation}

\paragraph{Global feature update.}
The global representation is updated by pooling node and edge features:
\begin{equation}
    y \leftarrow \mathrm{MLP}\bigl(y + \mathrm{Pool}(X) + \mathrm{Pool}(E)\bigr),
\end{equation}
where $\mathrm{Pool}$ denotes principal neighborhood aggregation (PNA) over nodes and edges respectively.

\subsection{FlexAttention Layer}
\label{app:flexatt}

\begin{figure}
    \centering
    \includegraphics[width=\textwidth]{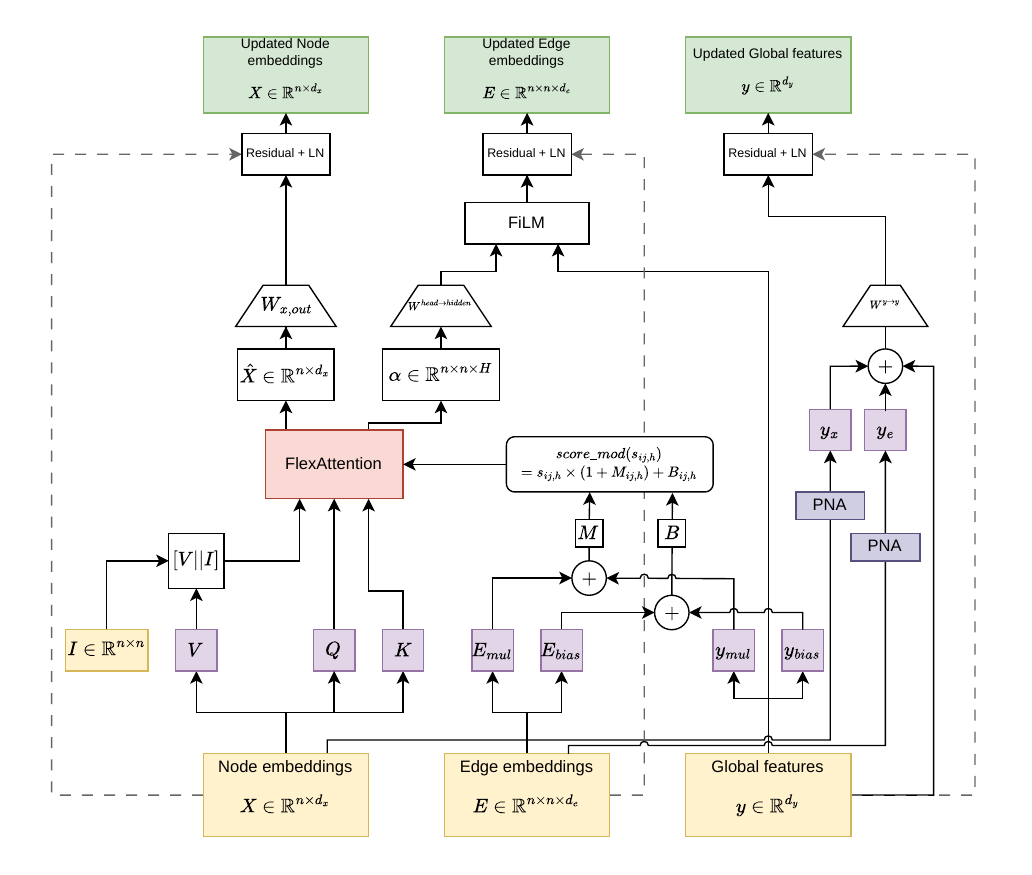}
    \caption{Schematic of the FlexAttentionGT architecture.}
    \label{fig:schematic_architecture}
\end{figure}

The FlexAttention layer preserves the three-stream structure of the original but replaces the element-wise outer product with standard dot-product attention, compresses the FiLM modulation to head-wise scalars, and recovers the attention map for the edge update in a single function call.
The overall architecture is illustrated in Figure~\ref{fig:schematic_architecture}.

\paragraph{Query, key, and value projections.}
Projections are identical to the original:
\begin{equation}
    Q, K, V = \mathrm{Linear}_{d_x}(X), \quad \text{each} \in \mathbb{R}^{N \times H \times d_f},
\end{equation}
then transposed to $\mathbb{R}^{H \times N \times d_f}$ as required by the FlexAttention interface.

\paragraph{Scalar dot-product attention logits.}
The unnormalized attention logit for head $h$ is a scalar:
\begin{equation}
    s_{ij,h} = \frac{q_{i,h} \cdot k_{j,h}}{\sqrt{d_f}} \in \mathbb{R}.
\end{equation}

\paragraph{Head-wise structural FiLM via \texttt{score\_mod}.}
Edge and global features are projected to head-wise scalar modifiers:
\begin{align}
    E_{\mathrm{mul}} &= \mathrm{Linear}_{H}(E) \in \mathbb{R}^{N \times N \times H}, \\
    E_{\mathrm{bias}} &= \mathrm{Linear}_{H}(E) \in \mathbb{R}^{N \times N \times H}, \\
    y_{\mathrm{mul}} &= \mathrm{Linear}_{H}(y) \in \mathbb{R}^{H}, \\
    y_{\mathrm{bias}} &= \mathrm{Linear}_{H}(y) \in \mathbb{R}^{H}.
\end{align}
These are combined into a multiplicative modifier $M \in \mathbb{R}^{N \times N \times H}$ and an additive bias $B \in \mathbb{R}^{N \times N \times H}$:
\begin{equation}
    M_{ij,h} = \sigma\bigl(E_{\mathrm{mul},ij,h} + y_{\mathrm{mul},h}\bigr), \quad
    B_{ij,h} = \sigma\bigl(E_{\mathrm{bias},ij,h} + y_{\mathrm{bias},h}\bigr),
\end{equation}
where $\sigma$ denotes the softsign function $\sigma(z) = z / (1 + |z|)$, used to bound the modifiers and stabilize training.
The modified score passed to the fused kernel is:
\begin{equation}
    s'_{ij,h} = \texttt{score\_mod}(s_{ij,h}) = s_{ij,h} \cdot (1 + M_{ij,h}) + B_{ij,h}.
    \label{eq:scoremod}
\end{equation}
Graph padding masks are handled via a \texttt{BlockMask}, which instructs the kernel to skip fully-masked blocks and enforces that $s'_{ij,h} \to -\infty$ for invalid pairs.

\paragraph{Single-pass node and edge update via $\bm{I_N}$ concatenation.}
FlexAttention does not materialize the softmax probabilities $\alpha_{ij,h}$ in global memory.
To recover them for the edge update without a second kernel pass, we concatenate the $N \times N$ identity matrix $I \in \mathbb{R}^{N \times N}$ to the value matrix along the feature axis:
\begin{equation}
    \mathrm{FlexAttn}\!\left(Q,\, K,\, [V \,\|\, I]\right)
    = \alpha\, [V \,\|\, I]
    = [\,\underbrace{\alpha V}_{\hat{X}} \,\|\, \underbrace{\alpha}_{\alpha}\,],
\end{equation}
where $\alpha \in \mathbb{R}^{N \times N \times H}$ are the softmax-normalized attention weights.
The first $d_f$ output channels give the aggregated node values $\hat{X} \in \mathbb{R}^{N \times d_x}$, and the trailing $N$ channels recover the full attention map $\alpha$.
Node and edge representations are thus both obtained from a single invocation of the fused kernel.

\paragraph{Edge update.}
The recovered attention map $\alpha \in \mathbb{R}^{N \times N \times H}$ is projected to the edge feature space and conditioned on global features:
\begin{equation}
    E'_{ij} = \mathrm{Linear}_{d_e}\!\left(\bigl(\mathrm{Linear}_{d_x}(y) + 1\bigr) \odot W_{\mathrm{head}} \,\alpha_{ij}\right) + \mathrm{Linear}_{d_e}(y) + E_{ij},
\end{equation}
where $W_{\mathrm{head}} \in \mathbb{R}^{d_x \times H}$ lifts the $H$ head scalars to $d_x$ before the global FiLM and final projection.
The two linear maps and the $y$-FiLM are fused algebraically into a single \texttt{einsum}, avoiding an explicit $N^2 \times d_x$ intermediate tensor.

\paragraph{Global feature update.}
Identical to the original layer:
\begin{equation}
    y \leftarrow \mathrm{MLP}\bigl(y + \mathrm{PNA}(X) + \mathrm{PNA}(E)\bigr).
\end{equation}

\paragraph{Comparison summary.}
Table~\ref{tab:layer_comparison} summarizes the key differences between the two layer variants.

\begin{table}[h]
\caption{Comparison of the original and FlexAttention layer variants.}
\label{tab:layer_comparison}
\centering
\small
\begin{tabular}{lll}
\toprule
& \textbf{GT (Original)} & \textbf{FlexAttentionGT (Ours)} \\
\midrule
Attention logit & $q_{i,h} \odot k_{j,h} \in \mathbb{R}^{d_f}$ & $q_{i,h} \cdot k_{j,h} \in \mathbb{R}$ \\
Softmax & Per-channel ($d_f$ distributions per $(i,h)$) & Standard (1 distribution per $(i,h)$) \\
Edge FiLM bandwidth & $d_x = H \cdot d_f$ channels & $H$ scalars \\
Edge FiLM location & Applied to $Y \in \mathbb{R}^{N \times N \times H \times d_f}$ & Inside fused kernel via \texttt{score\_mod} \\
Edge update source & Full $d_x$-dim bilinear $Y^{\mathrm{flat}}$ & $H$-dim attention map $\alpha$ \\
Node \& edge update & Two separate passes & Single fused kernel call \\
\bottomrule
\end{tabular}
\end{table}

\subsection{Further FlexAttention details}
\label{app:flexatt_details}

FlexAttention uses a FlashAttention-style \cite{dao2022flashattention} algorithm, tiling the query, key and value matrices and using the online softmax trick to never materialize the whole attention matrix at once which enables it to have a linear memory complexity with the number of queries/keys/values. The size of the tiling blocks and the specific backend that FlexAttention uses depends on the GPU type, available kernels and hidden dimensions of the query, key and value vectors. 

The tiled blocks of the query, key and value matrix have to fit into SRAM, which is limited and depends on the GPU. We have found good performance with blocks of size 32x32, but these parameters may need to be changed depending on the environment available and dataset used. Moreover, on recent GPUs, FlexAttention may select different kernels as a backend. Due to the identity matrix concatenation to the value matrix, its hidden dimension can be different from a power of 2 which kernels different from the default Triton one do not deal well with. We found more success forcing the backend to be Triton with the \texttt{kernel\_options} argument in FlexAttention.

\newpage

\section{Experiments Setup.}

\paragraph{Dataset size} We report the dataset size in \cref{tab:data_size}.

\begin{table}[ht]
    \caption{Sizes for different datasets we use in experiments.}
    \label{tab:data_size}
    \centering
    \begin{tabular}{lcc}
    \toprule
    Dataset & Min Nodes & Max Nodes  \\
    \midrule
    Planar & 64 & 64 \\
    Tree &  64 & 64 \\
    SBM  & 44 & 187  \\
    SBM-Large & 401 & 599 \\ 
    \bottomrule
    \end{tabular}
\end{table}

\paragraph{Sampling hyperparameters and training setup.}
We report in \cref{tab:training_sampling_hyperparam} the machine used for training, the number of training steps, and the sampling hyperparameters for the original GT architecture and our FlexAttention version. These settings correspond to the results in \cref{tab:generation_results} and the training curves in \cref{fig:gen_quality_curves}. For both the 50-step and 1000-step sampling curves, we evaluate checkpoints throughout training using the same sampling hyperparameters as in the official DeFoG implementation. All other training settings also follow the default configuration of the official DeFoG implementation. For the final reported results in Table \ref{tab:generation_results}, we select the best-performing hyperparameters for each model.

\begin{table}[ht]
\caption{Training and sampling settings for Planar, Tree, and SBM.}
\label{tab:training_sampling_hyperparam}
\centering
\small
\setlength{\tabcolsep}{4pt}
\begin{tabular}{lccccccc}
\toprule
Model 
& Initial Distribution 
& Train Distortion 
& Sample Distortion 
& $\omega$ 
& $\eta$ 
& GPU Type 
& Training steps \\
\midrule

\multicolumn{8}{c}{\textit{Planar}} \\
\midrule
GT (Original) & Marginal & Polydec & Polydec & 50 & 0.05 & V100 & 34000 \\
FlexAttentionGT (Ours) & Marginal & Polydec & Polydec & 50 & 0.02 & V100 & 34000 \\

\midrule
\multicolumn{8}{c}{\textit{Tree}} \\
\midrule
GT (Original) & Marginal & Polydec & Polydec & 0 & 0 & V100 & 26000 \\
FlexAttentionGT (Ours) & Marginal & Polydec & Polydec & 0 & 0 & V100 & 16500 \\

\midrule
\multicolumn{8}{c}{\textit{SBM}} \\
\midrule
GT (Original) & Absorbing & Identity & Identity & 0.0 & 0.0 & GH200 & 28000 \\
FlexAttentionGT (Ours) & Absorbing & Identity & Identity & 0.0 & 0.0 & GH200 & 28000 \\

\bottomrule
\end{tabular}
\end{table}

\end{document}

%% file: plot_inference.tex
\definecolor{nicepurple}{HTML}{8859AB}
\definecolor{nicegreen}{HTML}{88C2AB}
\definecolor{niceorange}{HTML}{CB722F}
\definecolor{niceblue}{HTML}{287DC5}
\definecolor{nicegray}{HTML}{363635}

\begin{tikzpicture}
    \begin{axis}[
        ybar,
        bar width=7pt,
        scale only axis,
        width=0.65\textwidth,
        height=3cm,
        enlarge x limits=0.25,
        enlarge y limits={upper,value=0.5},
        ylabel={Inference Speed (samples/s)},
        symbolic x coords={Planar, Tree, SBM},
        xtick=data,
        nodes near coords align={vertical},
        every node near coord/.append style={font=\tiny, rotate=90, anchor=west},
        ymin=0,
        ymajorgrids=true,
        grid style={dashed,gray!30},
    ]
    \addplot [fill=nicepurple, draw=none] table [col sep=comma, x=Dataset, y=Inference_Original] {efficiency_results.csv};

    \addplot [fill=nicegreen, draw=none, 
        visualization depends on={\thisrow{Inference_Original} \as \originalval},
        visualization depends on={\thisrow{Inference_Updated} \as \currval},
        nodes near coords={%
            \pgfmathparse{100*(\currval/\originalval - 1)}%
            \pgfmathprintnumber[fixed,precision=1,print sign]{\pgfmathresult}\%%
        }
    ] table [col sep=comma, x=Dataset, y=Inference_Updated] {efficiency_results.csv};
    \end{axis}
\end{tikzpicture}

%% file: plot_training.tex
\definecolor{nicepurple}{HTML}{8859AB}
\definecolor{nicegreen}{HTML}{88C2AB}
\definecolor{niceorange}{HTML}{CB722F}
\definecolor{niceblue}{HTML}{287DC5}
\definecolor{nicegray}{HTML}{363635}

\begin{tikzpicture}
    \begin{axis}[
        ybar,
        bar width=7pt,
        scale only axis,
        width=0.65\textwidth,
        height=3cm,
        enlarge x limits=0.25,
        enlarge y limits={upper,value=0.5},
        legend style={
            legend columns=-1,
            draw=none,
            fill=none,
            column sep=10pt
        },
        legend to name=sharedlegend,
        ylabel={Training Speed (samples/s)},
        symbolic x coords={Planar, Tree, SBM},
        xtick=data,
        nodes near coords align={vertical},
        every node near coord/.append style={font=\tiny, rotate=90, anchor=west},
        ymin=0,
        ymajorgrids=true,
        grid style={dashed,gray!30},
    ]
    \addplot [fill=nicepurple, draw=none] table [col sep=comma, x=Dataset, y=Training_Original] {efficiency_results.csv};

    \addplot [fill=nicegreen, draw=none, 
        visualization depends on={\thisrow{Training_Original} \as \originalval},
        visualization depends on={\thisrow{Training_Updated} \as \currval},
        nodes near coords={%
            \pgfmathparse{100*(\currval/\originalval - 1)}%
            \pgfmathprintnumber[fixed,precision=1,print sign]{\pgfmathresult}\%%
        }
    ] table [col sep=comma, x=Dataset, y=Training_Updated] {efficiency_results.csv};
    
    \legend{GT (Original), FlexAttentionGT (Ours)}
    \end{axis}
\end{tikzpicture}

%% file: plot_memory.tex
\definecolor{nicepurple}{HTML}{8859AB}
\definecolor{nicegreen}{HTML}{88C2AB}
\definecolor{niceorange}{HTML}{CB722F}
\definecolor{niceblue}{HTML}{287DC5}
\definecolor{nicegray}{HTML}{363635}

\begin{tikzpicture}
    \begin{axis}[
        ybar,
        bar width=7pt,
        scale only axis,
        width=0.65\textwidth,
        height=3cm,
        enlarge x limits=0.25,
        enlarge y limits={upper,value=0.1},
        ylabel={Memory (GB)},
        symbolic x coords={Planar, Tree, SBM},
        xtick=data,
        nodes near coords align={vertical},
        every node near coord/.append style={font=\tiny, rotate=90, anchor=west},
        ymin=0,
        ymajorgrids=true,
        grid style={dashed,gray!30},
    ]
    \addplot [fill=nicepurple, draw=none] table [col sep=comma, x=Dataset, y expr=\thisrow{Memory_Original}/1000] {efficiency_results.csv};

    \addplot [fill=nicegreen, draw=none, 
        visualization depends on={\thisrow{Memory_Original}/1000 \as \originalval},
        visualization depends on={\thisrow{Memory_Scaled}/1000 \as \currval}, 
        visualization depends on={\thisrow{Memory_Updated}/1000 \as \updval},
        nodes near coords={%
            \pgfmathparse{100*(\updval/\originalval - 1)}%
            \pgfmathprintnumber[fixed,precision=1,print sign]{\pgfmathresult}\%%
        }
    ] table [col sep=comma, x=Dataset, y expr=\thisrow{Memory_Updated}/1000] {efficiency_results.csv};
    \end{axis}
\end{tikzpicture}

%% file: plot_curve1000_planar.tex
\begin{tikzpicture}
\begin{axis}[
    width=\linewidth,
    height=3.8cm,
    xlabel={Training steps},
    ylabel={V.U.N.},
    ymin=0, ymax=1.05,
    enlarge x limits=0.05,
    xtick={0,10000,20000,30000},
    xticklabels={0,10k,20k,30k},
    scaled x ticks=false,
    ymajorgrids=true,
    grid style={dashed,gray!30},
    legend style={
        font=\small,
        draw=none,
        fill=none,
        at={(0.5,-0.3)},
        anchor=north,
        legend columns=2
    },
    legend to name=sharedlegend,
    legend columns=4,
    legend style={
        font=\small,
        draw=none,
        fill=none
    },
]

\addplot[color=nicepurple, mark=*]
table[col sep=comma, x=Steps, y=DeFoG] {vun_planar1000_4k.csv};
\addlegendentry{GT (Original)}

\addplot[color=niceblue, mark=square*]
table[col sep=comma, x=Steps, y=DeFoG_Small] {vun_planar1000_4k.csv};
\addlegendentry{GT (Scaled Down)}

\addplot[color=niceorange, mark=triangle*]
table[col sep=comma, x=Steps, y=DeFoG_LightEdge] {vun_planar1000_4k.csv};
\addlegendentry{GT (Light Edge)}

\addplot[color=nicegreen, mark=diamond*]
table[col sep=comma, x=Steps, y=Ours] {vun_planar1000_4k.csv};
\addlegendentry{FlexAttentionGT (Ours)}

\end{axis}
\end{tikzpicture}

%% file: plot_curve1000_sbm.tex
\begin{tikzpicture}
\begin{axis}[
    width=\linewidth,
    height=3.8cm,
    xlabel={Training steps},
    ylabel={V.U.N.},
    ymin=0, ymax=1.05,
    enlarge x limits=0.05,
    xtick={0,10000,20000,30000},
    xticklabels={0,10k,20k,30k},
    scaled x ticks=false,
    ymajorgrids=true,
    grid style={dashed,gray!30},
    legend style={
        font=\small,
        draw=none,
        fill=none,
        at={(0.5,-0.3)},
        anchor=north,
        legend columns=2
    },
    legend to name=sharedlegend,
    legend columns=4,
    legend style={
        font=\small,
        draw=none,
        fill=none
    },
]

\addplot[color=nicepurple, mark=*]
table[col sep=comma, x=Steps, y=DeFoG] {vun_sbm1000.csv};
\addlegendentry{GT (Original)}

\addplot[color=niceblue, mark=square*]
table[col sep=comma, x=Steps, y=DeFoG_Small] {vun_sbm1000.csv};
\addlegendentry{GT (Scaled Down)}

\addplot[color=niceorange, mark=triangle*]
table[col sep=comma, x=Steps, y=DeFoG_LightEdge] {vun_sbm1000.csv};
\addlegendentry{GT (Light Edge)}

\addplot[color=nicegreen, mark=diamond*]
table[col sep=comma, x=Steps, y=Ours] {vun_sbm1000.csv};
\addlegendentry{FlexAttentionGT (Ours)}

\end{axis}
\end{tikzpicture}

%% file: plot_curve50_planar.tex
\begin{tikzpicture}
\begin{axis}[
    width=\linewidth,
    height=3.8cm,
    xlabel={Training steps},
    ylabel={V.U.N.},
    ymin=0, ymax=1.05,
    enlarge x limits=0.05,
    xtick={0,10000,20000,30000},
    xticklabels={0,10k,20k,30k},
    scaled x ticks=false,
    ymajorgrids=true,
    grid style={dashed,gray!30},
    legend style={
        font=\small,
        draw=none,
        fill=none,
        at={(0.5,-0.25)},
        anchor=north,
        legend columns=2
    },
    legend to name=sharedlegend,
    legend columns=4,
    legend style={
        font=\small,
        draw=none,
        fill=none
    },
]

\addplot[color=nicepurple, mark=*]
table[col sep=comma, x=Steps, y=DeFoG] {vun_planar50_4k.csv};
\addlegendentry{GT (Original)}

\addplot[color=niceblue, mark=square*]
table[col sep=comma, x=Steps, y=DeFoG_Small] {vun_planar50_4k.csv};
\addlegendentry{GT (Scaled Down)}

\addplot[color=niceorange, mark=triangle*]
table[col sep=comma, x=Steps, y=DeFoG_LightEdge] {vun_planar50_4k.csv};
\addlegendentry{GT (Light Edge)}

\addplot[color=nicegreen, mark=diamond*]
table[col sep=comma, x=Steps, y=Ours] {vun_planar50_4k.csv};
\addlegendentry{FlexAttentionGT (Ours)}

\end{axis}
\end{tikzpicture}

%% file: plot_curve50_sbm.tex
\begin{tikzpicture}
\begin{axis}[
    width=\linewidth,
    height=3.8cm,
    xlabel={Training steps},
    ylabel={V.U.N.},
    ymin=0, ymax=1.05,
    enlarge x limits=0.05,
    xtick={0,10000,20000,30000},
    xticklabels={0,10k,20k,30k},
    scaled x ticks=false,
    ymajorgrids=true,
    grid style={dashed,gray!30},
    legend style={
        font=\small,
        draw=none,
        fill=none,
        at={(0.5,-0.25)},
        anchor=north,
        legend columns=2
    },
    legend to name=sharedlegend,
    legend columns=4,
    legend style={
        font=\small,
        draw=none,
        fill=none
    },
]

\addplot[color=nicepurple, mark=*]
table[col sep=comma, x=Steps, y=DeFoG] {vun_sbm50.csv};
\addlegendentry{GT (Original)}

\addplot[color=niceblue, mark=square*]
table[col sep=comma, x=Steps, y=DeFoG_Small] {vun_sbm50.csv};
\addlegendentry{GT (Scaled Down)}

\addplot[color=niceorange, mark=triangle*]
table[col sep=comma, x=Steps, y=DeFoG_LightEdge] {vun_sbm50.csv};
\addlegendentry{GT (Light Edge)}

\addplot[color=nicegreen, mark=diamond*]
table[col sep=comma, x=Steps, y=Ours] {vun_sbm50.csv};
\addlegendentry{FlexAttentionGT (Ours)}

\end{axis}
\end{tikzpicture}

%% file: plot_curve1000_planar_ablation_attncoeff.tex
\begin{tikzpicture}
\begin{axis}[
    width=\linewidth,
    height=3.8cm,
    xlabel={Training steps},
    ylabel={V.U.N.},
    ymin=0, ymax=1.05,
    enlarge x limits=0.05,
    xtick={0,10000,20000,30000},
    xticklabels={0,10k,20k,30k},
    scaled x ticks=false,
    ymajorgrids=true,
    grid style={dashed,gray!30},
    legend style={
        font=\small,
        draw=none,
        fill=none,
        at={(0.5,-0.25)},
        anchor=north,
        legend columns=2
    },
    legend to name=sharedlegend_ablation_edge,
    legend columns=2,
    legend style={
        font=\small,
        draw=none,
        fill=none
    },
]

\addplot[color=nicegreen, mark=diamond*]
table[col sep=comma, x=Steps, y=Ours] {vun_planar1000_ablation_attncoeff.csv};
\addlegendentry{FlexAttentionGT}

\addplot[color=nicegray, mark=triangle*, mark options={rotate=180}]
table[col sep=comma, x=Steps, y=Unnormalized] {vun_planar1000_ablation_attncoeff.csv};
\addlegendentry{FlexAttentionGT (Unnormalized)}

\end{axis}
\end{tikzpicture}

%% file: table_efficiency.tex
\begin{table}[ht]
\caption{Detailed computational efficiency results for the Planar, Tree, and SBM datasets. Inference and training speeds are measured in samples per second. Memory is reported during training, in MB. Tested on A100 GPU.}
\label{tab:efficiency_detailed}
\centering
\small
\begin{filecontents*}{efficiency_results.csv}
Dataset,Inference_Original,Inference_Scaled,Inference_Updated,Training_Original,Training_Scaled,Training_Updated,Memory_Original,Memory_Scaled,Memory_Updated
Planar,294.2,800.4,561.5,132.6,230.8,217.9,9579.6,2693.7,3323.4
Tree,310.6,775.5,549.1,147.7,417.6,217.7,9537.7,2651.7,3281.4
SBM,62.0,154.6,133.7,31.9,85.7,39.2,62839.9,21585.4,21991.4
\end{filecontents*}
\pgfplotstabletypeset[
    col sep=comma,
    every head row/.style={
        before row={
            \toprule
            & \multicolumn{2}{c}{Inference (samples/s) $\uparrow$} & \multicolumn{2}{c}{Training (samples/s) $\uparrow$} & \multicolumn{2}{c}{Memory (MB) $\downarrow$} \\
            \cmidrule(lr){2-3} \cmidrule(lr){4-5} \cmidrule(lr){6-7}
        },
        after row=\midrule
    },
    every last row/.style={after row=\bottomrule},
    columns={Dataset, Inference_Original, Inference_Updated, Training_Original, Training_Updated, Memory_Original, Memory_Updated},
    columns/Dataset/.style={string type, column type=l, column name=Dataset},
    columns/Inference_Original/.style={fixed, precision=1, column type=c, column name=GT},
    columns/Inference_Updated/.style={fixed, precision=1, column type=c, column name=FlexAttentionGT},
    columns/Training_Original/.style={fixed, precision=1, column type=c, column name=GT},
    columns/Training_Updated/.style={fixed, precision=1, column type=c, column name=FlexAttentionGT},
    columns/Memory_Original/.style={fixed, precision=0, column type=c, column name=GT},
    columns/Memory_Updated/.style={fixed, precision=0, column type=c, column name=FlexAttentionGT},
]{efficiency_results.csv}
\end{table}

%% file: table_efficiency_large_graphs.tex
\begin{table}[ht]
\caption{Efficiency benchmark on the SBM-Large dataset (batch size 2), with an average size of 500 nodes. Inference and training speeds are measured in samples per second. Memory is reported in MB. Tested on A100 GPU.}
\label{tab:efficiency_large_graphs}
\centering
\small
\begin{filecontents*}{efficiency_results_sbm500.csv}
Model,Forward_SPS,Forward_Mem,FwdBwd_SPS,FwdBwd_Mem
GT (Original),6.6,10412,3.4,43429
FlexAttentionGT (Single-pass),13.2,2584,0.9,15320
FlexAttentionGT (Two-pass),13.7,2595,6.1,15316
\end{filecontents*}
\pgfplotstabletypeset[
    col sep=comma,
    every head row/.style={
        before row={
            \toprule
            & \multicolumn{2}{c}{Inference} & \multicolumn{2}{c}{Training} \\
            \cmidrule(lr){2-3} \cmidrule(lr){4-5}
        },
        after row=\midrule
    },
    every last row/.style={after row=\bottomrule},
    columns={Model, Forward_SPS, Forward_Mem, FwdBwd_SPS, FwdBwd_Mem},
    columns/Model/.style={string type, column type=l, column name=Model},
    columns/Forward_SPS/.style={fixed, precision=1, column type=c, column name={Samples/s $\uparrow$}},
    columns/Forward_Mem/.style={fixed, precision=0, column type=c, column name={Mem (MB) $\downarrow$}},
    columns/FwdBwd_SPS/.style={fixed, precision=1, column type=c, column name={Samples/s $\uparrow$}},
    columns/FwdBwd_Mem/.style={fixed, precision=0, column type=c, column name={Mem (MB) $\downarrow$}},
]{efficiency_results_sbm500.csv}
\end{table}